\documentclass[11pt]{article}

\usepackage[preprint]{acl}

\usepackage{times}
\usepackage{latexsym}

\usepackage[T1]{fontenc}

\usepackage[utf8]{inputenc}

\usepackage{microtype}

\usepackage{inconsolata}
\usepackage{booktabs}
\usepackage{graphicx}
\usepackage[disable]{todonotes}
\usepackage{tipa}

\usepackage{algorithm}
\usepackage{algpseudocode}
\usepackage[table]{xcolor}

\newcommand{\ppllm}{\includegraphics[height=8pt,trim={100pt 50pt 100pt 30pt}, clip]{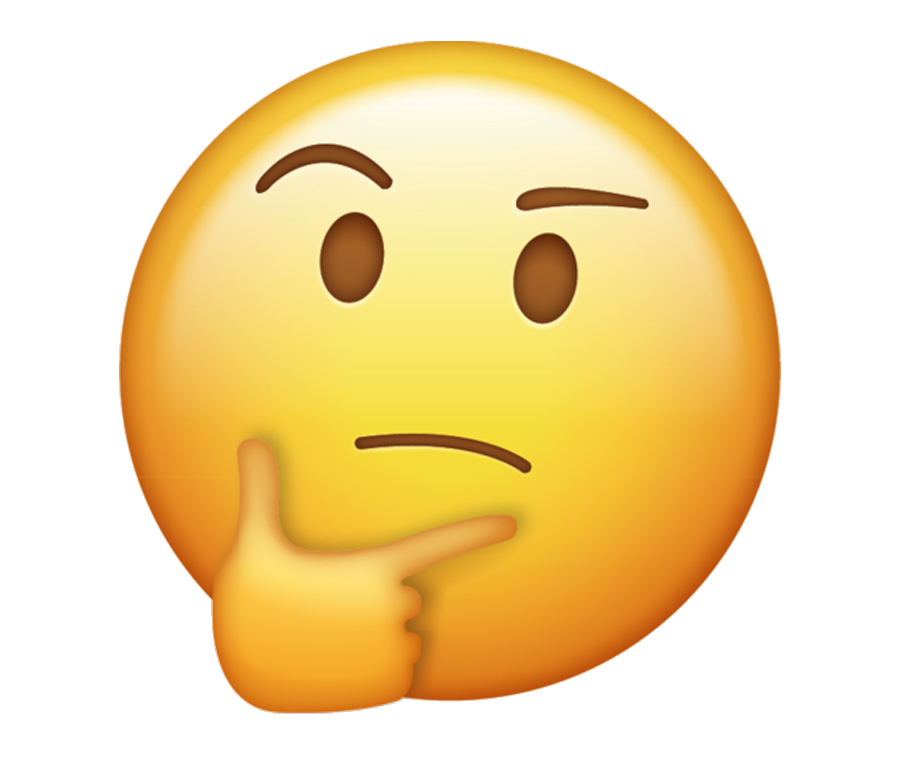}~\texttt{ppllm}}

\title{Large Language Models are Perplexed by some Political Parties}

\author{Paul Lerner \and François Yvon \\
        Sorbonne Université, CNRS, ISIR\\
        75005, Paris, France\\
        \texttt{lerner@isir.upmc.fr, yvon@isir.upmc.fr}
}

\begin{document}
\maketitle
\begin{abstract}
Large Language Models (LLMs) are increasingly used, including in political applications, but their political fairness has been little studied.
We assess it using perplexity, posing that a fair model should give equal probability to all political groups. 
However, we find, across ten
LLMs and three datasets covering 37 languages, that LLMs are more perplexed by the texts of  far right and nationalist parties than of social-democratic parties.
We find this to be consistent with previous work on translation fairness,  to the point that perplexity correlates with downstream translation metrics.
Our method is applicable to both base LLMs as well as their instruction-tuned counterpart, and we find that both are highly correlated, suggesting that  the political fairness of LLMs
stems from their pretraining, and is hardly affected
by instruction-tuning.

\end{abstract}

\section{Introduction}
\todo{rm TODOs}

Large Language Models (LLMs) are used daily by hundreds of millions of users through chatbots \citep{milmo2023chatgpt,similarweb} and are increasingly integrated in political applications \cite{DBLP:journals/corr/abs-2306-11932,tessler-etal-2024-ai,revel2025ai,polis2}.
We deem that LLMs should be \textit{fair} to everyone, regardless of their political beliefs.
\textit{Fairness} is most commonly defined for classification settings, through metrics such as equalized odds that enforce that every group is classified as accurately \citep{verma-rubin-2018-fairness,czarnowska-etal-2021-quantifying,barocas-etal-2023-fairness,gallegos-etal-2024-bias}.
Fairness of generative models has only begun to be studied quite recently, building upon the Computer Vision literature \cite{verine2026equalizedgenerativetreatmentmatching}.
As for the \textit{political fairness} of generative models, we are only aware of \citep{lerner2025assessingpoliticalfairnessmultilingual} who study how the quality of LLM-based translations vary according to the political affiliation of the speaker.

Whether they are used for summarization, translation, or plain chatting, the underlying computations performed by LLMs rely at their core on probability distributions defined over texts.
We therefore assess their political fairness using the well-known perplexity metric (along with other information-theoretic metrics). A fair model should give equal perplexity, i.e., average log-likelihood, to texts, irrespective of their partisan political content.
However, we find, across ten LLMs and three datasets covering 37 languages, %
that LLMs are more perplexed by the texts of some political parties than others. In particular, social-democratic parties consistently have a lower (i.e., better) perplexity than far left, far right and nationalist parties.

\begin{figure}[t]
    \centering
    \includegraphics[width=.9\linewidth,trim={3cm 0 0 0}, clip]{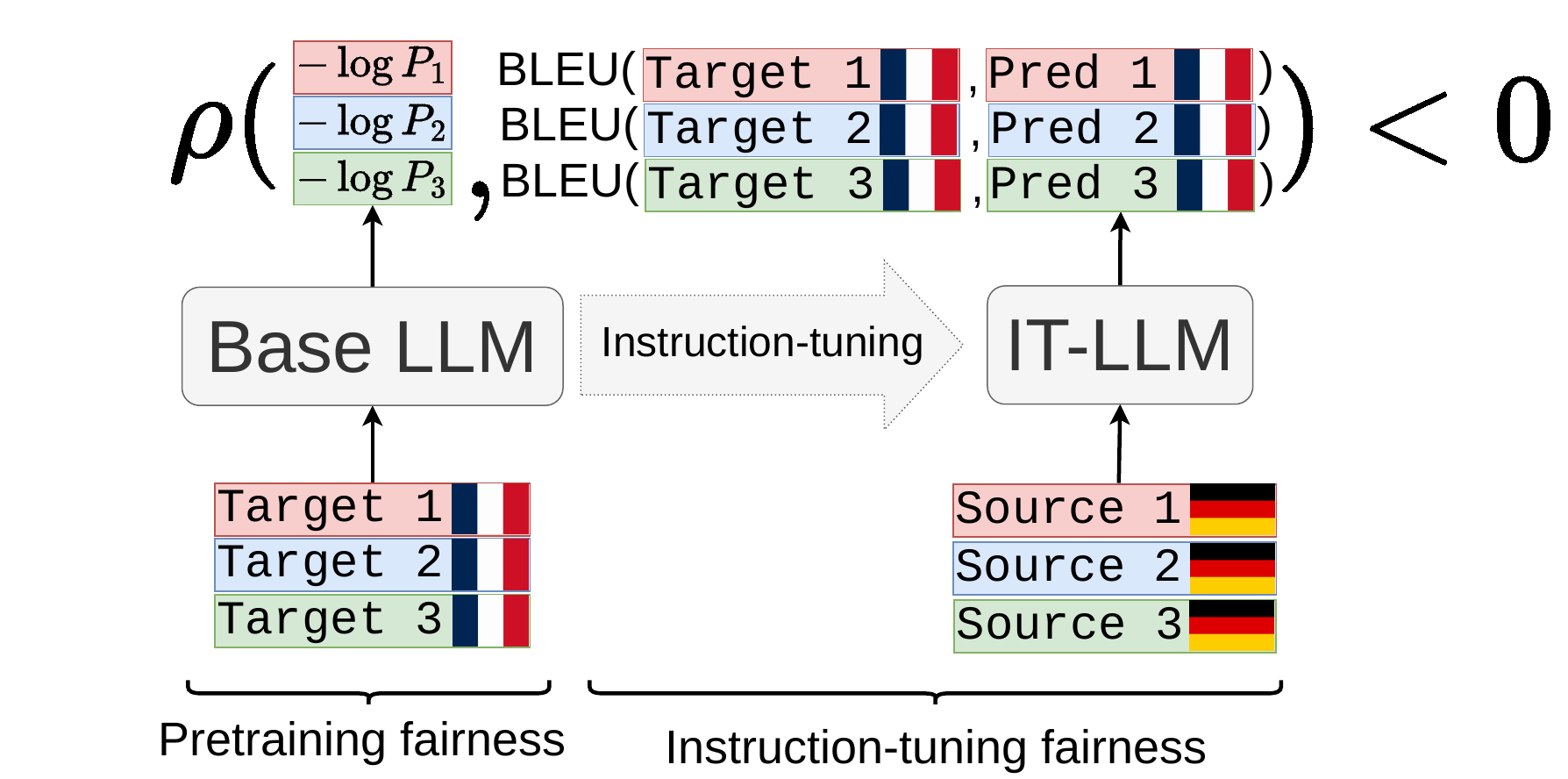}
    \caption{Simplified overview of our experiments. 
    Given texts from multiple political parties (different colors), we compute their negative log likelihood %
    using a base LLM. Differences in likelihood reveal unfair modeling and only rely on monolingual corpora. 
    We then compare parties likelihood with the downstream translation metrics (e.g., BLEU) of the IT-LLM counterpart, and find the correlation to be negative, showing the link between pretraining and instruction-tuning fairness.
    }
    \label{fig:base_v_itlm}
\end{figure}

Because this finding is consistent with that of \citet{lerner2025assessingpoliticalfairnessmultilingual},
we compute the  Spearman correlation between the per-party perplexities and downstream translation metrics  over 420 language pairs (sBLEU and COMET).
We find the correlation to be negative, as we expected since, by definition, LLMs are unlikely to generate a text which they assign little probability to.
Therefore, we can estimate the translation fairness of LLMs using plain monolingual corpora, without reference translations (as opposed to \citealp{lerner2025assessingpoliticalfairnessmultilingual}).
Furthermore, we find that the negative correlation holds even between the perplexity of the base LLM and the downstream translation metrics of its Instruction-tuned (IT-LLM) counterpart.
This result suggests that the political fairness of LLMs stems from their pretraining, and is hardly affected by  instruction-tuning.
This result calls for post-training alternative alignment methods, the design of fair pretraining methods, or a broader reflection on how we integrate LLMs in sensitive applications \citep{resnik-etal-2025-large}.
A sketch of these experiments is provided in Figure~\ref{fig:base_v_itlm}.

Our work sharply contrasts with the existing literature (Section~\ref{sec:rw}) that has mostly assessed the ``political biases'' of Instruction-tuned LLMs (IT-LLMs) by simulating their answers to English surveys \cite{feng-etal-2023-pretraining,rozado-etal-2023-political, santurkar-etal-2023-whose,motoki-etal-2024-more,potter-etal-2024-hidden,rottger-etal-2024-political,ceron-etal-2024-prompt,boelaert-etal-2024-how}. Our work follows near ten years of research on the gender and racial fairness of LLMs \cite{blodgett-etal-2020-language,gallegos-etal-2024-bias}. 

In summary, our main contribution is to assess the political fairness of LLMs through perplexity, which is applicable to both base and IT-LLMs on plain monolingual corpora and, as we found, predicts downstream translation fairness.

\section{Related Work}\label{sec:rw}
\subsection{Survey-based}
\paragraph{Early work} \citet{hartmann-etal-2023-political} evaluated ChatGPT soon after its release and found that its answers to German and Dutch voting advice applications aligned with the programs of socialists and green parties.

\paragraph{English} \citet{feng-etal-2023-pretraining,rozado-etal-2023-political, santurkar-etal-2023-whose,motoki-etal-2024-more,potter-etal-2024-hidden} have all surveyed multiple IT-LLMs using English surveys originally designed for (English-speaker) humans, including the Political Compass Test\footnote{\url{https://www.politicalcompass.org/}}, a popular test on the internet that gives scores along two axes depending on the agreement to statements such as ``\textit{Sex outside marriage is usually immoral}'': economic (left vs. right) and social (libertarian vs. authoritarian). These works overall concur that IT-LLMs are ``left-leaning''.

\paragraph{Brittleness} \citet{rottger-etal-2024-political,ceron-etal-2024-prompt,boelaert-etal-2024-how} %
have independently shown how brittle this survey method is, since the IT-LLM extracted answer varies depending on the exact phrasing of the prompt, e.g. using ``do you \emph{agree} or \emph{disagree}'' vs. ``do you \emph{disagree} or \emph{agree}'' may lead to different answers.

\paragraph{Multilingual} \citet{durmus-etal-2024-measuring,helwe-etal-2025-navigating} rely on Machine Translation (MT) to translate English surveys in multiple languages. However, this poses a confound since MT may (i) contain errors, including stance-changing errors \citep{nazanin}; 
(ii) be unfair to some political parties \citep{lerner2025assessingpoliticalfairnessmultilingual}.
Nevertheless, \citet{durmus-etal-2024-measuring} find that surveying IT-LLMs in English, Russian, Chinese, or Turkish always leads to occidental responses, i.e. the answers are consistent across languages. 
Likewise, \citet{weeber-etal-2026-political} find that aligning IT-LLMs in English transfers their political leaning to four other languages.
On the contrary, \citet{helwe-etal-2025-navigating} find that ``language has a strong influence'', although they do not control for the spurious prompt effects described above.
\citet{leo} have jointly assessed the variation caused by language and prompt phrasing. They found that (i) it was not possible to assess the bias of base LLMs through surveys, as the answer varied too much depending on the prompt phrasing; (ii) the effect of language was question-dependent. %

\paragraph{Pretraining vs. Instruction-tuning}
All works described in this section have only assessed the biases of IT-LLMs, with the exception of \citet{leo} who found that the survey method was not applicable to base LLMs.
However, instruction-tuning biases are easily explainable and controllable: it is trivial to fine-tune an LLM so that it gives the answer $y$ to the question $x$. 
On the contrary, pretraining biases seem much more deeply hidden. 

\subsection{Other Methods}
In addition to their survey-based method, \citet{potter-etal-2024-hidden} conduct a human-computer interaction experiment. They find that after debating with an IT-LLM in English, Trump supporters increased their Biden-leaning.

\citet{rottger-etal-2025-issuebench} assess the stance of IT-LLMs when prompted to generate English texts about debated issues. They find that LLMs align closely with US Democrats, as opposed to Republicans.

\citet{ceron-etal-2025-what} study the political content of LLMs English pretraining and posttraining data. Using a left-right content classifier, they find a majority of left-leaning data, which concurs with the results discussed above.

\citet{lerner2025assessingpoliticalfairnessmultilingual} study the MT fairness of IT-LLMs with respect to the political affiliation of the speaker in the European Parliament proceedings. %
They find that traditional left and right parties are favored over radical left and far right. However, the effect is not gradual since the green party is also poorly translated. 
We rely on their work since we compute the correlation between the perplexity of base LLMs and the downstream translation metrics of their IT-LLM counterpart. %

\subsection{Summary}
Despite methodological differences, all works point out that results are consistent across LLM families.
We differ from these works as we analyze the fairness of base LLMs and their IT-LLM counterpart by computing their perplexity on texts from various political parties, showing how some parties are unfairly modeled and how it relates to their unfair translations.

\section{Methods}
\subsection{Likelihood Metrics}\label{ssec:nll}
Given a text $t$ consisting of a sequence of $L$ tokens 
$(t_1,t_2,\dots,t_L)$, where $t_i$ is a symbol from a finite vocabulary (of e.g., words, subwords, or characters), a (large) language model defines a probability distribution $P(t)$, recursively factored as the product of $P(t_i|t_{<i})$ terms, which correspond to the likelihood of $t_i$ given its context $t_{<i}$ formed by the preceding tokens. The negative log-likelihood (NLL) of the entire sequence $t$ is defined as: 
\begin{equation}
    -\log_2 P(t) = \sum_{i=1}^L-\log_2 P(t_i|t_{<i}) 
    \label{eq:surprisal}
\end{equation}
with $t_0$, a special token that marks the beginning of the sequence.\footnote{Note that some models (e.g., Qwen3-8B) do not have such a beginning-of-sequence token. Therefore, the NLL sums from $i=2$, PPL is normalized by $L-1$, and we discount the characters in $t_1$ for BPC and BPEC.}
NLL (aka cross-entropy) is the loss that LLMs minimize during  pretraining. 
NLL is measured in bits and is interpreted as the amount of information needed to encode $t$ under $P$. 
If $t$ is the translation of a source text $s$, it is straightforward to compute its source-conditioned NLL $-\log_2 P(t|s)$ by concatenating $s$ to the context $t_{<i}$.\footnote{In practice, we use the same prompt as \citet{lerner2025assessingpoliticalfairnessmultilingual}: \texttt{"<src\_lang>: <src\_text>\textbackslash{}n<tgt\_lang>: <tgt\_text>"} to compute the source-conditioned NLL of the target text. For example: \texttt{"English: In Europe, we have freedom of the press.\textbackslash{}nFrench: En Europe la presse est libre."}}

NLL, sum of positive terms, increases along with $L$, i.e., longer texts are less likely than shorter texts. 
The most common way to account for that is to average the NLL over the $L$ tokens.
Exponentiating that gives rise to
perplexity (PPL), a standard measure to evaluate language models \citep{jm3}:
\begin{equation}
   \mathrm{PPL(t)} = 2^{-\frac{1}{L}\log_2 P(t)}
\end{equation}

However, because $L$ depends on the tokenizer, PPL is generally not comparable across models \citep{jm3}.
An attempt for a tokenizer-agnostic metric is to normalize NLL by the number of characters $\mathrm{C}(t)$ of the text, called 
bits per character (BPC, \citealp{sutskever-etal-2011-generating}):
\begin{equation}
   \mathrm{BPC(t)} = -\frac{1}{\mathrm{C}(t)}\log_2 P(t)
   \label{eq:bpc}
\end{equation}

However, as \citet{cotterell-etal-2018-are} argues, BPC is sensitive to orthographic artifacts (e.g., the word /put\textipa{S}/ is written in 3 characters `\textit{pu\v{c}}' in Czech  but in 6 characters `\textit{putsch}' in German). %
Provided with a parallel dataset, %
\citet{cotterell-etal-2018-are} introduce 
bits per English character (BPEC), which normalizes the NLL by the number of  characters $\mathrm{EC}(t)$ in the English translation of $t$:
\begin{equation}
   \mathrm{BPEC(t)} = -\frac{1}{\mathrm{EC}(t)}\log_2 P(t)
\end{equation}
We will present results using these four metrics.

\subsection{Political Fairness}\label{ssec:fairness_methods}
Given a monolingual corpus comprising texts of $K$ political parties, we compute PPL and BPC on each of the $K$ subsets to assess the political fairness of LLMs (both base and IT-LLMs).
To know whether the PPL and BPC significantly vary according to the party variable, we rely on a Kruskal-Wallis analysis, a non-parametric version of ANOVA \citep{kruskal1952use}. %

Given a multiparallel corpus, where each of the $N$ texts are available in $M$ languages, we compute BPEC in addition to PPL and BPC. We also compute the source-conditioned likelihood metrics for the $M(M-1)$ language pairs. These are applicable to both base and IT-LLMs.
To link pretraining fairness and downstream translation fairness, 
we finally compute the Spearman correlation between the likelihood metrics and the downstream translation metrics of IT-LLMs.
Following \citet{lerner2025assessingpoliticalfairnessmultilingual}, because automatic MT metrics sBLEU \citep{papineni-etal-2002-bleu,chen-cherry-2014-systematic} and COMET \citep{rei-etal-2022-comet22} are not directly comparable across languages, they are aggregated  per party  using Borda count  over all language pairs \citep{mclean2019a}.
Each language pair provides a ranking of $K$ parties based on their sBLEU (or COMET) score. 
The last-ranked party receives $0$ points, the second-to-last $1$ point, and so on up to the first-ranked party, which receives $K-1$ points. 
These scores are then averaged over all language pairs to obtain a final score.
An ideally fair model ranks each party at the same position in every language pair and thus achieves a score of $\frac{K-1}{2}$.

\subsection{Implementation}
Likelihood metrics are implemented with \ppllm{}.\footnote{\url{https://github.com/PaulLerner/ppllm}}

We experiment with pairs of base LLM and IT-LLM from  three families: Gemma-3-4B \citep{team-etal-2025-gemma}, Qwen3 \citep{yang-etal-2025-qwen3}, in two sizes (4B and 8B), and Llama-3.1 \citep{llamateam-etal-2024-llama}, in two sizes as well (8B and 70B). 

Experiments were carried out on a single V100, A100, or H100 GPU for all models except for Llama-3.1-70B which was run on four H100 GPUs. The experiments overall required approximately 2,500 GPU hours.

\section{Comparable Datasets}\label{sec:comparable}

\subsection{Manifesto and Parlamint}
\paragraph{Manifesto} Each language subset of Manifesto covers the programs, segmented in sentences, of tens of national parties \citep{merz-etal-2016-manifesto}. We use the texts from 2018 to today as test set.

\paragraph{Parlamint} Each language subset of Parlamint contains  debates from national parliaments, annotated with the party of the speaker \cite{erjavec-etal-2024-parlamint}. We use the texts from 2022 to today as test set.

\begin{figure}[t]
    \centering
    \includegraphics[width=\linewidth]{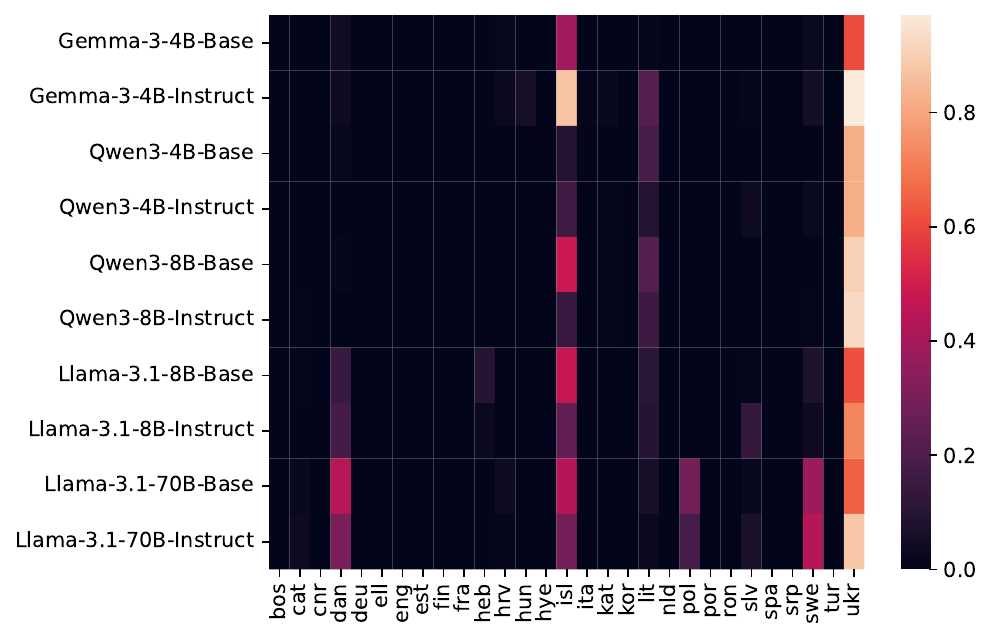}
    \caption{Kruskal-Wallis analysis of how BPC varies according to the party variable, showing the p-values across languages (columns) and models (lines) on the character-stratified subset of Manifesto. Dark colors show a small p-value, i.e., a statistically significant test depending on the chosen threshold.}
    \label{fig:ppllm_p_manifesto}
\end{figure}

\begin{figure}[t]
    \centering
    \includegraphics[width=\linewidth]{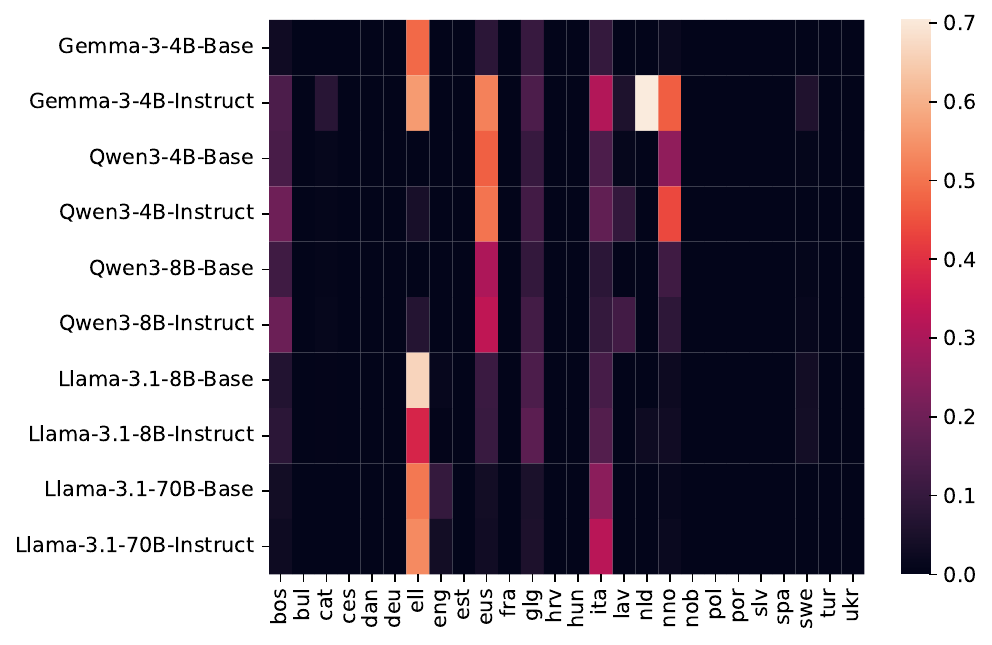}
    \caption{Kruskal-Wallis analysis of how BPC varies according to the party variable, showing the p-values across languages (columns) and models (lines) on the character-stratified subset of Parlamint. Dark colors show a small p-value, i.e., a statistically significant test depending on the chosen threshold.}
    \label{fig:ppllm_p_parlamint}
\end{figure}

\begin{table*}[t]
    \centering
    \resizebox{\textwidth}{!}{
    \begin{tabular}{rllrrrrrrrrr}
    \toprule
    Model &Stage& Dataset& BE & PCP & PS & SDP & L & PAN & CDS & IL &  CH\\
\midrule

    Gemma-3-4B & Base & Parlamint & \cellcolor{red!19} 667 & \cellcolor{green!30} 633 & \cellcolor{green!26} 636 & \cellcolor{green!22} 658 & \cellcolor{red!30} 687 & \cellcolor{green!18} 660 & \cellcolor{red!27} 686 & \cellcolor{red!15} 667 & \cellcolor{red!23} 684\\ 

Gemma-3-4B & Base & Manifesto & \cellcolor{red!27} 136 & \cellcolor{red!23} 135 & \cellcolor{green!26} 128 & \cellcolor{red!19} 134 & \cellcolor{green!18} 133 & \cellcolor{green!30} 126 & \cellcolor{green!22} 133 & \cellcolor{red!15} 134 & \cellcolor{red!30} 155\\ 

Gemma-3-4B & Instruct & Parlamint & \cellcolor{green!22} 901 & \cellcolor{green!30} 869 & \cellcolor{green!18} 902 & \cellcolor{red!19} 914 & \cellcolor{red!23} 923 & \cellcolor{green!26} 894 & \cellcolor{red!27} 924 & \cellcolor{red!15} 905 & \cellcolor{red!30} 940\\ 

Gemma-3-4B & Instruct & Manifesto & \cellcolor{red!27} 184 & \cellcolor{red!23} 184 & \cellcolor{green!26} 172 & \cellcolor{red!19} 181 & \cellcolor{green!22} 175 & \cellcolor{green!30} 172 & \cellcolor{red!15} 180 & \cellcolor{green!18} 179 & \cellcolor{red!30} 201\\ 

Qwen3-4B & Base & Parlamint & \cellcolor{red!15} 687 & \cellcolor{green!26} 657 & \cellcolor{green!30} 656 & \cellcolor{green!18} 681 & \cellcolor{red!27} 705 & \cellcolor{green!22} 680 & \cellcolor{red!23} 702 & \cellcolor{red!19} 691 & \cellcolor{red!30} 707\\ 

Qwen3-4B & Base & Manifesto & \cellcolor{red!27} 127 & \cellcolor{red!15} 125 & \cellcolor{green!26} 118 & \cellcolor{red!23} 126 & \cellcolor{green!22} 119 & \cellcolor{green!30} 116 & \cellcolor{green!18} 123 & \cellcolor{red!19} 126 & \cellcolor{red!30} 144\\ 
Qwen3-4B & Instruct & Parlamint & \cellcolor{red!15} 797 & \cellcolor{green!30} 760 & \cellcolor{green!26} 770 & \cellcolor{green!18} 794 & \cellcolor{red!27} 820 & \cellcolor{green!22} 791 & \cellcolor{red!23} 811 & \cellcolor{red!19} 801 & \cellcolor{red!30} 823\\ 

Qwen3-4B & Instruct & Manifesto & \cellcolor{red!27} 148 & \cellcolor{red!15} 144 & \cellcolor{green!22} 137 & \cellcolor{red!23} 148 & \cellcolor{green!26} 136 & \cellcolor{green!30} 136 & \cellcolor{green!18} 141 & \cellcolor{red!19} 145 & \cellcolor{red!30} 163\\ 

Qwen3-8B & Base & Parlamint & \cellcolor{red!15} 647 & \cellcolor{green!26} 617 & \cellcolor{green!30} 612 & \cellcolor{green!22} 637 & \cellcolor{red!30} 668 & \cellcolor{green!18} 642 & \cellcolor{red!23} 664 & \cellcolor{red!19} 650 & \cellcolor{red!27} 666\\ 

Qwen3-8B & Base & Manifesto & \cellcolor{red!27} 121 & \cellcolor{red!15} 119 & \cellcolor{green!26} 113 & \cellcolor{red!19} 120 & \cellcolor{green!22} 114 & \cellcolor{green!30} 111 & \cellcolor{green!18} 118 & \cellcolor{red!23} 121 & \cellcolor{red!30} 138\\ 

Qwen3-8B & Instruct & Parlamint & \cellcolor{red!15} 724 & \cellcolor{green!30} 687 & \cellcolor{green!26} 695 & \cellcolor{green!22} 717 & \cellcolor{red!30} 749 & \cellcolor{green!18} 721 & \cellcolor{red!23} 744 & \cellcolor{red!19} 727 & \cellcolor{red!27} 748\\ 

Qwen3-8B & Instruct & Manifesto & \cellcolor{red!27} 135 & \cellcolor{red!15} 130 & \cellcolor{green!26} 124 & \cellcolor{red!23} 134 & \cellcolor{green!22} 125 & \cellcolor{green!30} 124 & \cellcolor{green!18} 129 & \cellcolor{red!19} 134 & \cellcolor{red!30} 150\\ 

Llama-3.1-8B & Base & Parlamint & \cellcolor{red!15} 656 & \cellcolor{green!26} 622 & \cellcolor{green!30} 615 & \cellcolor{green!22} 643 & \cellcolor{red!30} 678 & \cellcolor{green!18} 649 & \cellcolor{red!23} 673 & \cellcolor{red!19} 660 & \cellcolor{red!27} 675\\ 

Llama-3.1-8B & Base & Manifesto & \cellcolor{red!19} 136 & \cellcolor{red!15} 134 & \cellcolor{green!26} 128 & \cellcolor{red!27} 137 & \cellcolor{green!22} 130 & \cellcolor{green!30} 127 & \cellcolor{green!18} 134 & \cellcolor{red!23} 137 & \cellcolor{red!30} 152\\ 

Llama-3.1-8B & Instruct & Parlamint & \cellcolor{red!15} 688 & \cellcolor{green!26} 651 & \cellcolor{green!30} 647 & \cellcolor{green!22} 674 & \cellcolor{red!30} 709 & \cellcolor{green!18} 680 & \cellcolor{red!27} 706 & \cellcolor{red!19} 690 & \cellcolor{red!23} 706\\ 

Llama-3.1-8B & Instruct & Manifesto & \cellcolor{red!23} 143 & \cellcolor{red!15} 142 & \cellcolor{green!22} 135 & \cellcolor{red!27} 144 & \cellcolor{green!26} 135 & \cellcolor{green!30} 132 & \cellcolor{green!18} 140 & \cellcolor{red!19} 143 & \cellcolor{red!30} 161\\ 

Llama-3.1-70B & Base & Parlamint & \cellcolor{red!15} 574 & \cellcolor{green!26} 539 & \cellcolor{green!30} 521 & \cellcolor{green!22} 556 & \cellcolor{red!30} 598 & \cellcolor{green!18} 570 & \cellcolor{red!27} 596 & \cellcolor{red!19} 579 & \cellcolor{red!23} 592\\ 

Llama-3.1-70B & Base & Manifesto & \cellcolor{red!19} 125 & \cellcolor{green!22} 120 & \cellcolor{green!26} 117 & \cellcolor{red!23} 126 & \cellcolor{green!18} 122 & \cellcolor{green!30} 113 & \cellcolor{red!15} 124 & \cellcolor{red!27} 127 & \cellcolor{red!30} 141\\ 

Llama-3.1-70B & Instruct & Parlamint & \cellcolor{red!15} 600 & \cellcolor{green!26} 564 & \cellcolor{green!30} 547 & \cellcolor{green!22} 581 & \cellcolor{red!30} 626 & \cellcolor{green!18} 596 & \cellcolor{red!27} 622 & \cellcolor{red!19} 605 & \cellcolor{red!23} 619\\ 

Llama-3.1-70B & Instruct & Manifesto & \cellcolor{red!23} 130 & \cellcolor{green!22} 124 & \cellcolor{green!26} 121 & \cellcolor{red!19} 130 & \cellcolor{green!18} 125 & \cellcolor{green!30} 119 & \cellcolor{red!15} 128 & \cellcolor{red!27} 132 & \cellcolor{red!30} 145\\

\bottomrule
    \end{tabular}}
    \caption{Average NLL ($\downarrow$) of Portuguese texts on Manifesto and Parlamint, in bits. Each dataset is separately stratified for the distribution of numbers of characters in the texts. Therefore, the numbers are not directly comparable across adjacent rows. Each row is colored from the highest (i.e. worst) NLL in red to the best (i.e. lowest) in green. Columns approximately position political parties from left to right. BE: Bloco de Esquerda; PCP: Partido Comunista Português; PS: Partido Socialista; SDP: Partido Social Democrata; L: Livre; PAN: Pessoas-Animais-Natureza; CDS: Centro Democrático Social - Partido Popular; IL: Iniciativa Liberal; CH: Chega.}
    \label{tab:pvm}
\end{table*}

\paragraph{Stratification}

We found that, for each language, the  parties texts had very different distributions in number of characters. That would confound our results since BPC (Equation~\ref{eq:bpc}) normalizes the NLL by the number of characters $\mathrm{C}(t)$ of a text $t$. 
While this normalization is supposed to balance the NLL formulation (Equation~\ref{eq:surprisal}), because the longer the context, the easier it is to predict the next token, we found a negative correlation between $\mathrm{BPC}(t)$ and $\mathrm{C}(t)$, across all models, languages, and both datasets.
To counter this confound, we stratified each language subset so that each party has a similar distribution in number of characters.
The target distribution is the minimum number of texts per bin over all parties. Therefore, we only considered parties with at least 100 texts. We use 100 bins with logarithmic scale.
The stratification algorithm is in Appendix~\ref{a:pm}. %
The following languages are therefore excluded because they have less than two parties with at least 100 texts each: \textsc{glg} and \textsc{lav} in Manifesto; \textsc{fin},  \textsc{isl},  \textsc{rus}, and  \textsc{srp} in  Parlamint.
The final number of parties and total number of texts for each language of both datasets are reported in Appendix~\ref{a:pm}. %

\subsection{Pretraining Fairness of National Parties}\label{ssec:pt_comp}%

Because Manifesto and Parlamint are not parallel corpora, only comparable ones, each of their language subset contains texts from tens of political parties, whose scores cannot be aggregated (e.g. \textit{Partido Socialista} and \textit{Partido Social Democrata} in \textsc{por} and \textit{La France insoumise} and \textit{Renaissance} in \textsc{fra}).
Therefore, to assess whether, in each language, some political parties are unfairly modeled, we conduct a Kruskal-Wallis analysis on the BPC of the character-stratified subsets of Manifesto and Parlamint.

The p-values of the Kruskal-Wallis tests are displayed in Figures~\ref{fig:ppllm_p_manifesto} and \ref{fig:ppllm_p_parlamint} for Manifesto and Parlamint, respectively. 
We find that across datasets and models, most languages BPC variance is significantly explained by the party variable ($p<0.01$). 
This means that the texts of some political parties are unfairly modeled by LLMs. 
We find this result to be consistent across model training stages (Base or IT), sizes, and even families.

The few languages for which the Kruskal-Wallis test is often not statistically significant ($p>0.01$) across models have little data available after stratification: only 63 texts per party for \textsc{isl} and \textsc{ukr} in Manifesto, and less than 150 texts per party for \textsc{ell}, \textsc{eus}, \textsc{ita}, and \textsc{nno} in Parlamint (see Appendix~\ref{a:pm}).

Of course, it does not provide much insight to know that political parties are unfairly modeled in most languages and both datasets. We seek to know \textit{which} parties are \textit{poorly} modeled, and which are \textit{well} modeled. %
This is why we devote the next section to Portuguese (\textsc{por}), which is well-covered by both Manifesto and Parlamint.

\subsection{A case study of Portuguese}\label{ssec:por} %

Manifesto and Parlamint cover hundreds of texts for each of the nine Portuguese political parties (Table~\ref{tab:pvm}).
Because each dataset was stratified so that its political parties' texts have a comparable distribution of numbers of characters, their NLL is comparable and we report it in Table~\ref{tab:pvm} (as it is more readable than BPC). %

We find that the speeches from radical-left BE and far-right CH have the highest (i.e. worst) NLL, on the character-stratified subsets of both Manifesto and Parlamint. 
On the opposite, the speeches from the left PS have the lowest (i.e. best) NLL.
These findings are consistent with those of \citet{lerner2025assessingpoliticalfairnessmultilingual} for Machine Translation.
Again, these findings are consistent across model training stages
(Base or IT), sizes, and families.

\subsection{Pretraining Fairness of Party Families}\label{ssec:fam}

\begin{table}[t]
    \centering
    \resizebox{\columnwidth}{!}{
    \begin{tabular}{rlrrrrr}
    \toprule
        Model & Stage & LEF & SOC & LIB & CON & NAT \\ 
        \midrule 
        Fair Model & --&2.0&2.0&2.0&2.0&2.0\\
        \midrule
Gemma-3-4B&Base & 2.0 & 3.3 & 1.7 & 2.5 & 0.5 \\
Gemma-3-4B&Instruct & 1.6 & 3.0 & 1.8 & 3.1 & 0.5 \\
Qwen3-4B&Base & 1.9 & 3.0 & 1.8 & 2.6 & 0.7 \\
Qwen3-4B&Instruct & 1.5 & 3.3 & 1.9 & 2.7 & 0.6 \\
Qwen3-8B&Base & 1.8 & 3.0 & 1.9 & 2.6 & 0.7 \\
Qwen3-8B&Instruct & 1.9 & 3.2 & 2.0 & 2.5 & 0.4 \\
Llama-3.1-8B&Base & 2.2 & 3.0 & 1.6 & 2.5 & 0.7 \\
Llama-3.1-8B&Instruct & 2.0 & 3.0 & 1.5 & 2.7 & 0.8 \\
Llama-3.1-70B&Base & 2.1 & 3.3 & 1.8 & 2.4 & 0.4 \\
Llama-3.1-70B&Instruct & 2.1 & 3.1 & 1.7 & 2.5 & 0.6 \\

        \bottomrule
    \end{tabular}}
    \caption{Borda Count of political party families based on BPC (from 0-4, the higher the better), averaged over 10 languages in Manifesto.         LEF: Socialist or other left parties, SOC: Social democratic parties, LIB: Liberal parties,        CON: Conservative parties, NAT: Nationalist and radical right parties.}
    \label{tab:Manifesto}
\end{table}

While the previous section focused on the national parties of a single country, here we broaden our scope by aggregating the metrics of national parties belonging to the same family (from left to nationalist, see Table~\ref{tab:Manifesto}).
However, Manifesto is not a parallel corpus, only a comparable one. 
Therefore, we cannot directly compute the BPC of a given party family, say CON (conservative) parties, by concatenating its manifests in every language, as party families are not equally represented in every language.
Instead, we leverage the same Borda-based ranking aggregation as described in Section~\ref{ssec:fairness_methods} as
each language provides a ranking of the $K$ party families based on their BPC (the lower, the better).
Aiming at maximizing $K$, the number of party families considered, as well as the number of voting languages, we restrict to the following 10 languages that cover each of the $K=5$ largest party families in Manifesto (listed in Table~\ref{tab:Manifesto}) after character stratification: \textsc{dan}, \textsc{deu}, \textsc{eng}, \textsc{fin}, \textsc{heb}, \textsc{ita}, \textsc{por}, \textsc{slv}, \textsc{spa}, and \textsc{swe}. %

Results on the character-stratified subset of Manifesto are reported in Table~\ref{tab:Manifesto}.
Consistently with results reported above, %
we find that NAT (nationalist and radical right) parties have most often the worst BPC while  SOC (social democratic) parties have often the best BPC.
The consistency across languages is striking: e.g., for Llama-3.1-70B-Base, NAT have a score of 0.4 (i.e., almost always ranked last across languages) while SOC have a score of 3.2 (i.e., almost always ranked first across languages).
Further analysis is impeded as we do not have access to the training data of these models, we do not even know on which languages they were trained.
Again, the results are largely consistent across model sizes, training stages, and families.

\begin{figure}[t]
    \centering
    \includegraphics[width=.8\linewidth]{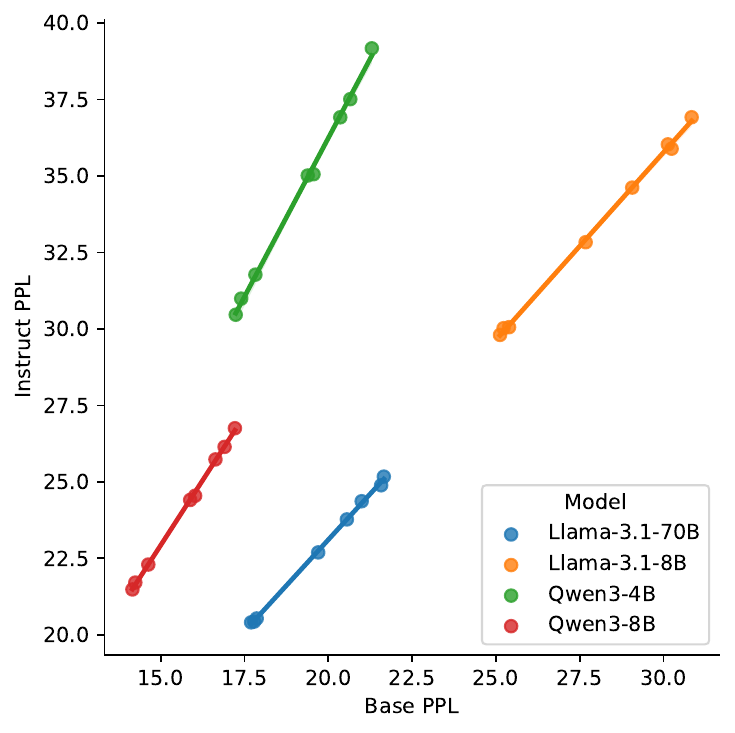}
    \caption{PPL ($\downarrow$) of base LLMs plotted against IT-LLMs. For each model, each point represents  a political party of 21-EuroParl, and the correlation reaches a Pearson $r$ of .99. PPL is computed on the target text $t$, without context, especially without the source text. PPL is \textit{not} comparable across model families (e.g., between Qwen3 and Llama-3.1).}
    \label{fig:ppl}
\end{figure}

\begin{table*}[t]
\centering
    \begin{tabular}{rlrrrrrrrr}
    \toprule
Model & Stage& NGL & S\&D & EFA & ALDE & EPP & ECR & EFD & NA\\
\midrule

Gemma-3-4B & Base & \cellcolor{green!17} 41.0 & \cellcolor{green!25} 36.5 & \cellcolor{red!22} 43.5 & \cellcolor{red!18} 42.2 & \cellcolor{green!30} 36.5 & \cellcolor{red!30} 45.1 & \cellcolor{green!21} 37.6 & \cellcolor{red!26} 45.1\\ 

Gemma-3-4B & Instruct & \cellcolor{green!17} 174.0 & \cellcolor{green!25} 157.3 & \cellcolor{red!26} 198.6 & \cellcolor{red!18} 182.9 & \cellcolor{green!30} 153.1 & \cellcolor{red!22} 197.2 & \cellcolor{green!21} 158.5 & \cellcolor{red!30} 210.5\\ 

Qwen3-4B & Base & \cellcolor{red!18} 19.6 & \cellcolor{green!25} 17.4 & \cellcolor{red!22} 20.4 & \cellcolor{green!17} 19.4 & \cellcolor{green!30} 17.2 & \cellcolor{red!26} 20.7 & \cellcolor{green!21} 17.8 & \cellcolor{red!30} 21.3\\ 

Qwen3-4B & Instruct & \cellcolor{red!18} 35.1 & \cellcolor{green!25} 31.0 & \cellcolor{red!22} 36.9 & \cellcolor{green!17} 35.0 & \cellcolor{green!30} 30.5 & \cellcolor{red!26} 37.5 & \cellcolor{green!21} 31.8 & \cellcolor{red!30} 39.2\\ 

Qwen3-8B & Base & \cellcolor{red!18} 16.0 & \cellcolor{green!25} 14.2 & \cellcolor{red!22} 16.6 & \cellcolor{green!17} 15.9 & \cellcolor{green!30} 14.2 & \cellcolor{red!26} 16.9 & \cellcolor{green!21} 14.6 & \cellcolor{red!30} 17.2\\ 

Qwen3-8B & Instruct & \cellcolor{red!18} 24.5 & \cellcolor{green!25} 21.7 & \cellcolor{red!22} 25.7 & \cellcolor{green!17} 24.4 & \cellcolor{green!30} 21.5 & \cellcolor{red!26} 26.1 & \cellcolor{green!21} 22.3 & \cellcolor{red!30} 26.8\\ 

Llama-3.1-8B & Base & \cellcolor{green!17} 27.7 & \cellcolor{green!25} 25.2 & \cellcolor{red!22} 30.1 & \cellcolor{red!18} 29.1 & \cellcolor{green!30} 25.1 & \cellcolor{red!26} 30.2 & \cellcolor{green!21} 25.4 & \cellcolor{red!30} 30.8\\ 

Llama-3.1-8B & Instruct & \cellcolor{green!17} 32.8 & \cellcolor{green!25} 30.0 & \cellcolor{red!26} 36.0 & \cellcolor{red!18} 34.6 & \cellcolor{green!30} 29.8 & \cellcolor{red!22} 35.9 & \cellcolor{green!21} 30.1 & \cellcolor{red!30} 36.9\\ 

Llama-3.1-70B & Base & \cellcolor{green!17} 19.7 & \cellcolor{green!30} 17.7 & \cellcolor{red!22} 21.0 & \cellcolor{red!18} 20.6 & \cellcolor{green!25} 17.8 & \cellcolor{red!26} 21.6 & \cellcolor{green!21} 17.9 & \cellcolor{red!30} 21.7\\ 

Llama-3.1-70B & Instruct & \cellcolor{green!17} 22.7 & \cellcolor{green!25} 20.4 & \cellcolor{red!22} 24.4 & \cellcolor{red!18} 23.8 & \cellcolor{green!30} 20.4 & \cellcolor{red!26} 24.9 & \cellcolor{green!21} 20.5 & \cellcolor{red!30} 25.2\\ 
\bottomrule
\end{tabular}

    \caption{PPL ($\downarrow$) of base and IT-LLMs for the eight political parties of 21-EuroParl. PPL is computed on the target text $t$, without context, especially without the source text. PPL is \textit{not} comparable across model families (e.g., between Gemma-3 and Qwen3). Each row is colored from the highest (i.e. worst) PPL in red to the best (i.e. lowest) in green. Columns approximately position political parties from left to right. NGL: European United Left/Nordic Green Left; S\&D: Progressive Alliance of Socialists and Democrats; EFA: Group of the Greens/European Free Alliance; ALDE: Group of the Alliance of Liberals and Democrats for Europe; EPP: Group of the European People's Party; ECR: European Conservatives and Reformists Group; EFD: Europe of freedom and democracy Group;    NA: Non-attached Members.}
    \label{tab:ppl}
\end{table*}

\section{Parallel Dataset}
\subsection{21-EuroParl}
21-EuroParl \citep{lerner2025assessingpoliticalfairnessmultilingual} is a multiparallel version of the proceedings of the European Parliament from years 2009-2011, aligned at the sentence level. The dataset comprises 72,234 examples, where each example is a sentence translated in $M=21$ languages: \textsc{bul}, \textsc{ces}, \textsc{dan}, \textsc{deu}, \textsc{ell}, \textsc{eng}, \textsc{est}, \textsc{fin}, \textsc{fra}, \textsc{hun}, \textsc{ita}, \textsc{lav}, \textsc{lit}, \textsc{nld}, \textsc{pol}, \textsc{por}, \textsc{ron}, \textsc{slk}, \textsc{slv}, \textsc{spa}, and \textsc{swe} (ISO 639-3).\footnote{\textsc{slk} is the 37\textsuperscript{th} language we study in the paper, the other being already covered either in Manifesto or Parlamint.}
Each example is annotated with extensive political metadata, including the European Party of the speaker. There are $K=8$ parties (see Table~\ref{tab:ppl}). We conduct all experiments on the test set of $N=23386$ examples.

\subsection{Pretraining Fairness on a Parallel Dataset}\label{ssec:pt_parallel}%

Table~\ref{tab:ppl} reports the PPL of base and IT-LLMs for the 8 political parties of 21-EuroParl. %
Party rankings based on  base LLMs PPL are consistent with those of their IT-LLM counterpart used for MT:
EPP and S\&D get the best (i.e. lowest) PPL, while smaller parties like NA get the worst (i.e. highest) PPL, consistently across all model families and sizes.
The correlation between the PPL of base LLMs and their IT-LLM counterpart is shown  in Figure~\ref{fig:ppl} (Gemma is excluded from the figure to improve readability, because of its high PPL, as reported in Table~\ref{tab:ppl}, but also reaches a high correlation of Pearson $r=.97$). %
Appendix~\ref{sa:pt_metrics} also reports BPC and BPEC, showing consistent results.

As \citet{lerner2025assessingpoliticalfairnessmultilingual}, we do not find a gradual political preference (e.g., from left to right) but rather that some political parties, namely the traditional left S\&D and right  EPP, are favored over smaller parties, from radical left (NGL) and far right (NA) but also center (EFA and ALDE).

\subsection{Pretraining Fairness correlates with Downstream Performance}\label{ssec:corr}

Table~\ref{tab:cond_ppl_v_trad} reports the Spearman $\rho$  correlation between source-conditioned PPL and sBLEU- and COMET-based Borda  counts across political parties in 21-EuroParl.\footnote{The Borda counts are taken from \citet[Table 6] {lerner2025assessingpoliticalfairnessmultilingual}.} 
The correlation is negative, both for base LLMs and  their IT-LLM counterpart, 
i.e., when a given model assigns little probability to the reference translation of a given political party, 
it is unlikely to generate it, which leads to a poor sBLEU (and COMET) score.
The result is consistent for both metrics but the correlation is stronger with COMET.
The result also holds with source-conditioned BPC and BPEC (see Appendix~\ref{a:metrics}). 

Table~\ref{tab:bpec_v_trad} extends this analysis by showing the Spearman $\rho$  correlation between  unconditioned PPL of the target text and sBLEU- and COMET-based Borda counts across political parties.
The correlation, while smaller, is also negative.
That means that when a given model assigns little probability to the text of a given political party, even outside of any translation context, 
it is unlikely to generate it, including in a translation context, which leads to  a poor sBLEU (or COMET) score.
Again, the result is consistent for both metrics but the correlation is stronger with COMET.
Again, the same holds for  BPC and BPEC (see Appendix~\ref{a:metrics}).

For a given LLM, the correlation of the Base stage is very close, if not equal, to that of the IT stage.
This can be explained by what we have seen previously: for a given LLM, the PPL of the Base stage is nearly perfectly correlated to that of the IT stage. 
Again, the same holds for  BPC and BPEC.

\begin{table}[t]
    \centering
    \resizebox{\columnwidth}{!}{
    \begin{tabular}{rlrr}
\toprule
    \textbf{Model} & \textbf{Stage}&\textbf{sBLEU} & \textbf{COMET} \\
\midrule
            Gemma-3-4B&Base & -0.690 & -0.690 \\ 
        Gemma-3-4B&Instruct & -0.524 & -0.690 \\ 
        Qwen3-4B&Base & -0.833 & -0.952 \\ 
        Qwen3-4B&Instruct & -0.643 & -0.905 \\ 
        Qwen3-8B&Base & -0.833 & -0.857 \\ 
        Qwen3-8B&Instruct & -0.690 & -0.833 \\ 
        Llama-3.1-8B&Base & -0.571 & -0.738 \\ 
        Llama-3.1-8B&Instruct & -0.667 & -0.833 \\ 
        Llama-3.1-70B&Base & -0.671 & -0.643 \\ 
        Llama-3.1-70B&Instruct & -0.635 & -0.714 \\ 
        \bottomrule
    \end{tabular}}
    \caption{Spearman $\rho$ shows a negative correlation between source-conditioned PPL ($\downarrow$) and sBLEU- and COMET-based Borda counts ($\uparrow$) across political parties, both for the source-conditioned PPL of base LLMs  and of IT-LLMs. The translations are always generated by the IT-LLM.}
    \label{tab:cond_ppl_v_trad}
\end{table}

\begin{table}[t]
    \centering
    \resizebox{\columnwidth}{!}{
    \begin{tabular}{rlrr}
\toprule
    \textbf{Model}  & \textbf{Stage}& \textbf{sBLEU} & \textbf{COMET} \\
\midrule
        Gemma-3-4B&Base & -0.262 & -0.429 \\ 
        Gemma-3-4B&Instruct & -0.357 & -0.548 \\ 
        Qwen3-4B&Base & -0.667 & -0.881 \\ 
        Qwen3-4B&Instruct & -0.667 & -0.881 \\ 
        Qwen3-8B&Base & -0.738 & -0.810 \\ 
        Qwen3-8B&Instruct & -0.738 & -0.810 \\ 
        Llama-3.1-8B&Base & -0.405 & -0.548 \\ 
        Llama-3.1-8B&Instruct & -0.429 & -0.619 \\ 
        Llama-3.1-70B&Base & -0.527 & -0.524 \\ 
        Llama-3.1-70B&Instruct & -0.527 & -0.524 \\ 
        \bottomrule
    \end{tabular}}
    \caption{Spearman $\rho$ shows a negative correlation between PPL ($\downarrow$) and sBLEU- and COMET-based Borda counts ($\uparrow$) across political parties, both for the PPL of base LLMs  and of IT-LLMs in 21-EuroParl. PPL is computed on the target text $t$, without context, especially without the source text. The translations are always generated by the IT-LLM.}
    \label{tab:bpec_v_trad}
\end{table}

\section{Discussion}

We found that the PPL, BPC, and BPEC of a given base LLM  are all negatively correlated with the downstream translation metrics of its IT-LLM counterpart, on the texts of political parties  (Section~\ref{ssec:corr}).
In other words, when a base LLM assigns a small probability to the text of a given political party, it will be unlikely to generate it, so the automatic translation into other languages will be also poor, even after going through instruction-tuning. 
These results suggest that the political fairness of LLMs stems from their pretraining, and that it is hardly affected by instruction-tuning. 
This finding calls for alternative alignment methods, the design of fair pretraining methods, or a broader reflection on how we integrate LLMs in sensitive applications \citep{resnik-etal-2025-large}.
Future work may extend the political analysis of pretraining data of \citet{ceron-etal-2025-what} to more languages than English and finer-grained labels than left-wing vs. right-wing. The same can be said about the interpretability study of \citet{kim2025linear}.

Equipped with this correlation finding, we can take a look back to the results on comparable datasets that cover more languages (Section~\ref{sec:comparable}).
Indeed, we found that
(i) for the large majority of the 36 languages studied, BPC significantly varies according to the party variable (Section~\ref{ssec:pt_comp}), so we can expect the translation quality to also vary;
(ii) in Portuguese, the speeches from the left PS are better modeled than those of the radical-left BE and far-right CH (Section~\ref{ssec:por}), so we can expect  that the speeches of PS will be better translated than those of BE and CH;
(iii) across 10 languages, social-democratic parties are better modeled than far right and nationalist parties (Section~\ref{ssec:fam}), so we expect they will be better translated as well.
Compared to the MT study of \citet{lerner2025assessingpoliticalfairnessmultilingual}, we cover more languages and datasets, since we can apply our method to plain monolingual corpora.

Our work goes beyond the left-right dichotomy of the survey-based related work (Section~\ref{sec:rw}) and provides evidence that the political fairness of LLMs is not gradual.

Finally, multilingualism is core to our work as our analysis was carried out over 36 languages (Section~\ref{sec:comparable}) and 420 language pairs (Section~\ref{ssec:corr}).
Understanding how languages are entangled in a single model is a fundamental question for multilingual NLP. Several works point towards the influence of English (as the majority language in the training data) and to language-independent features in the middle layers of the model
 \cite{wendler-etal-2024-llamas,guo-etal-2025-large,wang-etal-2025-lost-multilinguality}. 
We found that the political fairness of a given LLM was very consistent across languages. This suggests that the political features of texts are shared across all languages within the model, as opposed to language-specific subspaces. 
This argues in favor of viewing multilingual LLMs as ``polyglots'' having  consistent evaluation scores across the languages they model.
We deem that this finding provides ample avenues for future work.

\section*{Limitations}
Our study is limited to LLMs of up to 70 billion parameters, as we could not afford to run the largest models, e.g. the 235B Qwen3 or 405B Llama-3.1. However, our experiment with Llama-3.1 shows that, although the performance overall improves from 8B to 70B, the political fairness stays consistent.

As often in NLP, the datasets used in our experiments may have leaked into the training sets of LLMs.

Our study is limited to texts of politicians in formal settings: either political programs or parliamentary proceedings of speeches. Our findings may differ in more informal settings (e.g., social media).

\section*{Ethical considerations}
Our analysis is based upon the political affiliation of parliamentarians or political programs, both of which are of public interest. 
We do not foresee any ethical issues with our work.

\section*{Acknowledgments}
This research was funded by BPI-France under the project AI For Democracy - Democratic Commons, one of seven winners of BPI-France's ``Digital Commons for Generative AI'' call for projects, conducted as part of the France 2030 investment plan.
We thank all the members of the Democratic Commons program at Sorbonne Université, Sciences Po, and make.org.

\bibliography{custom}

\begin{thebibliography}{47}
\providecommand{\natexlab}[1]{#1}

\bibitem[{Barocas et~al.(2023)Barocas, Hardt, and
  Narayanan}]{barocas-etal-2023-fairness}
Solon Barocas, Moritz Hardt, and Arvind Narayanan. 2023.
\newblock \emph{Fairness and {{Machine Learning}}: {{Limitations}} and
  {{Opportunities}}}.
\newblock MIT Press.

\bibitem[{Blodgett et~al.(2020)Blodgett, Barocas, Daum{\'e}~III, and
  Wallach}]{blodgett-etal-2020-language}
Su~Lin Blodgett, Solon Barocas, Hal Daum{\'e}~III, and Hanna Wallach. 2020.
\newblock \href {https://doi.org/10.18653/v1/2020.acl-main.485} {Language
  ({{Technology}}) is {{Power}}: {{A Critical Survey}} of ``{{Bias}}'' in
  {{NLP}}}.
\newblock In \emph{Proceedings of the 58th {{Annual Meeting}} of the
  {{Association}} for {{Computational Linguistics}}}, pages 5454--5476, Online.
  Association for Computational Linguistics.

\bibitem[{Boelaert et~al.(2025)Boelaert, Coavoux, Étienne Ollion, Petev, and
  Präg}]{boelaert-etal-2024-how}
Julien Boelaert, Samuel Coavoux, Étienne Ollion, Ivaylo Petev, and Patrick
  Präg. 2025.
\newblock \href {https://doi.org/10.1177/00491241251330582} {Machine bias. how
  do generative language models answer opinion polls?1}.
\newblock \emph{Sociological Methods \& Research}, 54(3):1156--1196.

\bibitem[{Ceron et~al.(2024)Ceron, Falk, Bari{\'c}, Nikolaev, and
  Pad{\'o}}]{ceron-etal-2024-prompt}
Tanise Ceron, Neele Falk, Ana Bari{\'c}, Dmitry Nikolaev, and Sebastian
  Pad{\'o}. 2024.
\newblock \href {https://doi.org/10.1162/tacl_a_00710} {Beyond {{Prompt
  Brittleness}}: {{Evaluating}} the {{Reliability}} and {{Consistency}} of
  {{Political Worldviews}} in {{LLMs}}}.
\newblock \emph{Transactions of the Association for Computational Linguistics},
  12:1378--1400.

\bibitem[{Ceron et~al.(2025)Ceron, Nikolaev, Stammbach, and
  Nozza}]{ceron-etal-2025-what}
Tanise Ceron, Dmitry Nikolaev, Dominik Stammbach, and Debora Nozza. 2025.
\newblock What is the political content in {LLMs'} pre-and post-training data?
\newblock \emph{arXiv preprint arXiv:2509.22367}.

\bibitem[{Chen and Cherry(2014)}]{chen-cherry-2014-systematic}
Boxing Chen and Colin Cherry. 2014.
\newblock \href {https://doi.org/10.3115/v1/W14-3346} {A systematic comparison
  of smoothing techniques for sentence-level {BLEU}}.
\newblock In \emph{Proceedings of the Ninth Workshop on Statistical Machine
  Translation}, pages 362--367, Baltimore, Maryland, USA. Association for
  Computational Linguistics.

\bibitem[{Computational Democracy~Project(2026)}]{polis2}
The Computational Democracy~Project. 2026.
\newblock \href
  {https://web.archive.org/web/20260216174419/https://pol.is/home2} {Polis
  2.0}.
\newblock Polis. Accessed on March 11th 2026. URL archived from February 21st
  2026.

\bibitem[{Cotterell et~al.(2018)Cotterell, Mielke, Eisner, and
  Roark}]{cotterell-etal-2018-are}
Ryan Cotterell, Sabrina~J. Mielke, Jason Eisner, and Brian Roark. 2018.
\newblock \href {https://doi.org/10.18653/v1/N18-2085} {Are {{All Languages
  Equally Hard}} to {{Language-Model}}?}
\newblock In \emph{Proceedings of the 2018 {{Conference}} of the {{North
  American Chapter}} of the {{Association}} for {{Computational Linguistics}}:
  {{Human Language Technologies}}, {{Volume}} 2 ({{Short Papers}})}, pages
  536--541, New Orleans, Louisiana. Association for Computational Linguistics.

\bibitem[{Czarnowska et~al.(2021)Czarnowska, Vyas, and
  Shah}]{czarnowska-etal-2021-quantifying}
Paula Czarnowska, Yogarshi Vyas, and Kashif Shah. 2021.
\newblock \href {https://doi.org/10.1162/tacl_a_00425} {Quantifying social
  biases in {NLP}: {A} generalization and empirical comparison of extrinsic
  fairness metrics}.
\newblock \emph{Transactions of the Association for Computational Linguistics},
  9:1249--1267.
\newblock Place: Cambridge, MA Publisher: MIT Press.

\bibitem[{Durmus et~al.(2024)Durmus, Nguyen, Liao, Schiefer, Askell, Bakhtin,
  Chen, Hatfield-Dodds, Hernandez, Joseph, Lovitt, McCandlish, Sikder, Tamkin,
  Thamkul, Kaplan, Clark, and Ganguli}]{durmus-etal-2024-measuring}
Esin Durmus, Karina Nguyen, Thomas Liao, Nicholas Schiefer, Amanda Askell,
  Anton Bakhtin, Carol Chen, Zac Hatfield-Dodds, Danny Hernandez, Nicholas
  Joseph, Liane Lovitt, Sam McCandlish, Orowa Sikder, Alex Tamkin, Janel
  Thamkul, Jared Kaplan, Jack Clark, and Deep Ganguli. 2024.
\newblock \href {https://openreview.net/forum?id=zl16jLb91v} {Towards measuring
  the representation of subjective global opinions in language models}.
\newblock In \emph{First Conference on Language Modeling}.

\bibitem[{Erjavec et~al.(2024)Erjavec, Kopp, Ljube{\v s}i{\'c}, Kuzman, Rayson,
  Osenova, Ogrodniczuk, {\c C}{\"o}ltekin, Kor{\v z}inek, Meden, Skubic,
  Rupnik, Agnoloni, Aires, Barkarson, Bartolini, Bel, Calzada~P{\'e}rez, Dar{\c
  g}is, Diwersy, Gavriilidou, {van Heusden}, Iruskieta, Kahusk, Kryvenko,
  {Ligeti-Nagy}, Magari{\~n}os, M{\"o}lder, Navarretta, Simov, Tungland,
  Tuominen, Vidler, Vladu, Wissik, Yrj{\"a}n{\"a}inen, and Fi{\v
  s}er}]{erjavec-etal-2024-parlamint}
Toma{\v z} Erjavec, Maty{\'a}{\v s} Kopp, Nikola Ljube{\v s}i{\'c}, Taja
  Kuzman, Paul Rayson, Petya Osenova, Maciej Ogrodniczuk, {\c C}a{\u g}r{\i}
  {\c C}{\"o}ltekin, Danijel Kor{\v z}inek, Katja Meden, Jure Skubic, Peter
  Rupnik, Tommaso Agnoloni, Jos{\'e} Aires, Starka{\dh}ur Barkarson, Roberto
  Bartolini, N{\'u}ria Bel, Mar{\'i}a Calzada~P{\'e}rez, Roberts Dar{\c g}is,
  and 18 others. 2024.
\newblock \href {https://doi.org/10.1007/s10579-024-09798-w} {{{ParlaMint II}}:
  Advancing comparable parliamentary corpora across {{Europe}}}.
\newblock \emph{Language Resources and Evaluation}.

\bibitem[{Feng et~al.(2023)Feng, Park, Liu, and
  Tsvetkov}]{feng-etal-2023-pretraining}
Shangbin Feng, Chan~Young Park, Yuhan Liu, and Yulia Tsvetkov. 2023.
\newblock \href {https://doi.org/10.18653/v1/2023.acl-long.656} {From
  pretraining data to language models to downstream tasks: Tracking the trails
  of political biases leading to unfair {NLP} models}.
\newblock In \emph{Proceedings of the 61st Annual Meeting of the Association
  for Computational Linguistics (Volume 1: Long Papers)}, pages 11737--11762,
  Toronto, Canada. Association for Computational Linguistics.

\bibitem[{Gallegos et~al.(2024)Gallegos, Rossi, Barrow, Tanjim, Kim,
  Dernoncourt, Yu, Zhang, and Ahmed}]{gallegos-etal-2024-bias}
Isabel~O. Gallegos, Ryan~A. Rossi, Joe Barrow, Md~Mehrab Tanjim, Sungchul Kim,
  Franck Dernoncourt, Tong Yu, Ruiyi Zhang, and Nesreen~K. Ahmed. 2024.
\newblock \href {https://doi.org/10.1162/coli_a_00524} {Bias and {{Fairness}}
  in {{Large Language Models}}: {{A Survey}}}.
\newblock \emph{Computational Linguistics}, 50(3):1097--1179.

\bibitem[{Grattafiori et~al.(2024)Grattafiori, Dubey, Jauhri, Pandey, Kadian,
  Al-Dahle, Letman, Mathur, Schelten, Vaughan
  et~al.}]{llamateam-etal-2024-llama}
Aaron Grattafiori, Abhimanyu Dubey, Abhinav Jauhri, Abhinav Pandey, Abhishek
  Kadian, Ahmad Al-Dahle, Aiesha Letman, Akhil Mathur, Alan Schelten, Alex
  Vaughan, and 1 others. 2024.
\newblock The llama 3 herd of models.
\newblock \emph{arXiv preprint arXiv:2407.21783}.

\bibitem[{Guo et~al.(2025)Guo, Conia, Zhou, Li, Potdar, and
  Xiao}]{guo-etal-2025-large}
Yanzhu Guo, Simone Conia, Zelin Zhou, Min Li, Saloni Potdar, and Henry Xiao.
  2025.
\newblock \href {https://doi.org/10.18653/v1/2025.acl-long.193} {Do large
  language models have an {E}nglish accent? evaluating and improving the
  naturalness of multilingual {LLM}s}.
\newblock In \emph{Proceedings of the 63rd Annual Meeting of the Association
  for Computational Linguistics (Volume 1: Long Papers)}, pages 3823--3838,
  Vienna, Austria. Association for Computational Linguistics.

\bibitem[{Hartmann et~al.(2023)Hartmann, Schwenzow, and
  Witte}]{hartmann-etal-2023-political}
Jochen Hartmann, Jasper Schwenzow, and Maximilian Witte. 2023.
\newblock \href {https://doi.org/10.2139/ssrn.4316084} {The political ideology
  of conversational ai: Converging evidence on chatgpt’s pro-environmental,
  left-libertarian orientation}.
\newblock \emph{Available at SSRN 4316084}.

\bibitem[{Helwe et~al.(2025)Helwe, Balalau, and
  Ceolin}]{helwe-etal-2025-navigating}
Chadi Helwe, Oana Balalau, and Davide Ceolin. 2025.
\newblock \href {https://doi.org/10.18653/v1/2025.findings-acl.883} {Navigating
  the {{Political Compass}}: {{Evaluating Multilingual LLMs}} across
  {{Languages}} and {{Nationalities}}}.
\newblock In \emph{Findings of the {{Association}} for {{Computational
  Linguistics}}: {{ACL}} 2025}, pages 17179--17204, Vienna, Austria.
  Association for Computational Linguistics.

\bibitem[{Jurafsky and Martin(2026)}]{jm3}
Daniel Jurafsky and James~H. Martin. 2026.
\newblock \href {https://web.stanford.edu/~jurafsky/slp3/} {\emph{Speech and
  Language Processing: An Introduction to Natural Language Processing,
  Computational Linguistics, and Speech Recognition, with Language Models}},
  3rd edition.
\newblock Online manuscript released January 6, 2026.

\bibitem[{Kim et~al.(2025)Kim, Evans, and Schein}]{kim2025linear}
Junsol Kim, James Evans, and Aaron Schein. 2025.
\newblock \href {https://openreview.net/forum?id=rwqShzb9li} {Linear
  representations of political perspective emerge in large language models}.
\newblock In \emph{The Thirteenth International Conference on Learning
  Representations}.

\bibitem[{Kruskal and Wallis(1952)}]{kruskal1952use}
William~H Kruskal and W~Allen Wallis. 1952.
\newblock Use of ranks in one-criterion variance analysis.
\newblock \emph{Journal of the American statistical Association},
  47(260):583--621.

\bibitem[{Labat et~al.(2026)Labat, Ollion, and Yvon}]{leo}
Léo Labat, Etienne Ollion, and François Yvon. 2026.
\newblock \href {https://arxiv.org/abs/2602.05932} {Polyglots or multitudes?
  multilingual llm answers to value-laden multiple-choice questions}.
\newblock \emph{Preprint}, arXiv:2602.05932.

\bibitem[{Lerner and
  Yvon(2025)}]{lerner2025assessingpoliticalfairnessmultilingual}
Paul Lerner and François Yvon. 2025.
\newblock \href {https://arxiv.org/abs/2510.20508} {Assessing the political
  fairness of multilingual llms: A case study based on a 21-way multiparallel
  europarl dataset}.
\newblock \emph{Preprint}, arXiv:2510.20508.

\bibitem[{McLean(2019)}]{mclean2019a}
I~McLean. 2019.
\newblock \emph{Voting}, page 121–140.
\newblock Oxford University Press.

\bibitem[{Merz et~al.(2016)Merz, Regel, and
  Lewandowski}]{merz-etal-2016-manifesto}
Nicolas Merz, Sven Regel, and Jirka Lewandowski. 2016.
\newblock \href {https://doi.org/10.1177/2053168016643346} {The {{Manifesto
  Corpus}}: {{A}} new resource for research on political parties and
  quantitative text analysis}.
\newblock \emph{Research \& Politics}, 3(2):2053168016643346.

\bibitem[{Milmo et~al.(2023)}]{milmo2023chatgpt}
Dan Milmo and 1 others. 2023.
\newblock Chatgpt reaches 100 million users two months after launch.
\newblock \emph{The Guardian}, 3:1017--1054.

\bibitem[{Motoki et~al.(2024)Motoki, Pinho~Neto, and
  Rodrigues}]{motoki-etal-2024-more}
Fabio Motoki, Valdemar Pinho~Neto, and Victor Rodrigues. 2024.
\newblock \href {https://doi.org/10.1007/s11127-023-01097-2} {More human than
  human: Measuring {{ChatGPT}} political bias}.
\newblock \emph{Public Choice}, 198(1):3--23.

\bibitem[{Papineni et~al.(2002)Papineni, Roukos, Ward, and
  Zhu}]{papineni-etal-2002-bleu}
Kishore Papineni, Salim Roukos, Todd Ward, and Wei-Jing Zhu. 2002.
\newblock Bleu: A method for automatic evaluation of machine translation.
\newblock In \emph{Proceedings of the 40th Annual Meeting of the
  {{Association}} for {{Computational Linguistics}}}, pages 311--318.

\bibitem[{Potter et~al.(2024)Potter, Lai, Kim, Evans, and
  Song}]{potter-etal-2024-hidden}
Yujin Potter, Shiyang Lai, Junsol Kim, James Evans, and Dawn Song. 2024.
\newblock \href {https://doi.org/10.18653/v1/2024.emnlp-main.244} {Hidden
  {{Persuaders}}: {{LLMs}}' {{Political Leaning}} and {{Their Influence}} on
  {{Voters}}}.
\newblock In \emph{Proceedings of the 2024 {{Conference}} on {{Empirical
  Methods}} in {{Natural Language Processing}}}, pages 4244--4275, Miami,
  Florida, USA. Association for Computational Linguistics.

\bibitem[{Rei et~al.(2022)Rei, {C. de Souza}, Alves, Zerva, Farinha, Glushkova,
  Lavie, Coheur, and Martins}]{rei-etal-2022-comet22}
Ricardo Rei, Jos{\'e}~G. {C. de Souza}, Duarte Alves, Chrysoula Zerva, Ana~C
  Farinha, Taisiya Glushkova, Alon Lavie, Luisa Coheur, and Andr{\'e} F.~T.
  Martins. 2022.
\newblock {{COMET-22}}: {{Unbabel-IST}} 2022 {{Submission}} for the {{Metrics
  Shared Task}}.
\newblock In \emph{Proceedings of the {{Seventh Conference}} on {{Machine
  Translation}} ({{WMT}})}, pages 578--585, Abu Dhabi, United Arab Emirates
  (Hybrid). Association for Computational Linguistics.

\bibitem[{Resnik(2025)}]{resnik-etal-2025-large}
Philip Resnik. 2025.
\newblock \href {https://doi.org/10.1162/coli_a_00558} {Large {{Language Models
  Are Biased Because They Are Large Language Models}}}.
\newblock \emph{Computational Linguistics}, 51(3):885--906.

\bibitem[{Revel and Penigaud(2025)}]{revel2025ai}
Manon Revel and Theophile Penigaud. 2025.
\newblock Ai-facilitated collective judgements.
\newblock \emph{Available at SSRN 5167340}.

\bibitem[{R{\"o}ttger et~al.(2025)R{\"o}ttger, Hinck, Hofmann, Hackenburg,
  Pyatkin, Brahman, and Hovy}]{rottger-etal-2025-issuebench}
Paul R{\"o}ttger, Musashi Hinck, Valentin Hofmann, Kobi Hackenburg, Valentina
  Pyatkin, Faeze Brahman, and Dirk Hovy. 2025.
\newblock Issuebench: Millions of realistic prompts for measuring issue bias in
  {LLM} writing assistance.
\newblock \emph{arXiv preprint arXiv:2502.08395}.

\bibitem[{R{\"o}ttger et~al.(2024)R{\"o}ttger, Hofmann, Pyatkin, Hinck, Kirk,
  Schuetze, and Hovy}]{rottger-etal-2024-political}
Paul R{\"o}ttger, Valentin Hofmann, Valentina Pyatkin, Musashi Hinck, Hannah
  Kirk, Hinrich Schuetze, and Dirk Hovy. 2024.
\newblock \href {https://doi.org/10.18653/v1/2024.acl-long.816} {Political
  compass or spinning arrow? {Towards} more meaningful evaluations for values
  and opinions in large language models}.
\newblock In \emph{Proceedings of the 62nd Annual Meeting of the Association
  for Computational Linguistics (Volume 1: Long Papers)}, pages 15295--15311,
  Bangkok, Thailand. Association for Computational Linguistics.

\bibitem[{Rozado(2023)}]{rozado-etal-2023-political}
David Rozado. 2023.
\newblock \href {https://doi.org/10.3390/socsci12030148} {The {{Political
  Biases}} of {{ChatGPT}}}.
\newblock \emph{Social Sciences}, 12(3):148.

\bibitem[{Santurkar et~al.(2023)Santurkar, Durmus, Ladhak, Lee, Liang, and
  Hashimoto}]{santurkar-etal-2023-whose}
Shibani Santurkar, Esin Durmus, Faisal Ladhak, Cinoo Lee, Percy Liang, and
  Tatsunori Hashimoto. 2023.
\newblock Whose {{Opinions Do Language Models Reflect}}?
\newblock In \emph{Proceedings of the 40th {{International Conference}} on
  {{Machine Learning}}}, pages 29971--30004. PMLR.

\bibitem[{Shafiabadi and Yvon(2026)}]{nazanin}
Nazanin Shafiabadi and François Yvon. 2026.
\newblock \href {https://doi.org/10.63317/2pjio9ho8rxg} {Biases in translation:
  Assessing opinion distortion in machine translated texts}.
\newblock In \emph{Proceedings of the Fifteenth Language Resources and
  Evaluation Conference (LREC 2026)}, pages 8596--8614, Palma, Mallorca, Spain.
  European Language Resources Association (ELRA).

\bibitem[{Small et~al.(2023)Small, Vendrov, Durmus, Homaei, Barry, Cornebise,
  Suzman, Ganguli, and Megill}]{DBLP:journals/corr/abs-2306-11932}
Christopher~T. Small, Ivan Vendrov, Esin Durmus, Hadjar Homaei, Elizabeth
  Barry, Julien Cornebise, Ted Suzman, Deep Ganguli, and Colin Megill. 2023.
\newblock \href {https://doi.org/10.48550/arXiv.2306.11932} {Opportunities and
  risks of llms for scalable deliberation with polis}.
\newblock \emph{CoRR}, abs/2306.11932.

\bibitem[{Sutskever et~al.(2011)Sutskever, Martens, and
  Hinton}]{sutskever-etal-2011-generating}
Ilya Sutskever, James Martens, and Geoffrey Hinton. 2011.
\newblock Generating text with recurrent neural networks.
\newblock In \emph{Proceedings of the 28th {{International Conference}} on
  {{International Conference}} on {{Machine Learning}}}, {{ICML}}'11, pages
  1017--1024, Madison, WI, USA. Omnipress.

\bibitem[{Team(2025)}]{team-etal-2025-gemma}
Gemma Team. 2025.
\newblock \href {https://doi.org/10.48550/arXiv.2503.19786} {Gemma 3
  {{Technical Report}}}.
\newblock \emph{Preprint}, arXiv:2503.19786.

\bibitem[{Tessler et~al.(2024)Tessler, Bakker, Jarrett, Sheahan, Chadwick,
  Koster, Evans, {Campbell-Gillingham}, Collins, Parkes, Botvinick, and
  Summerfield}]{tessler-etal-2024-ai}
Michael~Henry Tessler, Michiel~A. Bakker, Daniel Jarrett, Hannah Sheahan,
  Martin~J. Chadwick, Raphael Koster, Georgina Evans, Lucy
  {Campbell-Gillingham}, Tantum Collins, David~C. Parkes, Matthew Botvinick,
  and Christopher Summerfield. 2024.
\newblock \href {https://doi.org/10.1126/science.adq2852} {{{AI}} can help
  humans find common ground in democratic deliberation}.
\newblock \emph{Science}, 386(6719):eadq2852.

\bibitem[{Top Websites~Ranking(2025)}]{similarweb}
The Top Websites~Ranking. 2025.
\newblock \href
  {https://web.archive.org/web/20251010190024/https://www.similarweb.com/top-websites/}
  {The top 50 most visited websites for september 2025}.
\newblock Similarweb. Accessed on October 13th 2025. URL archived from October
  10th 2025.

\bibitem[{Verine et~al.(2026)Verine, Pinot, and
  Bronnec}]{verine2026equalizedgenerativetreatmentmatching}
Alexandre Verine, Rafael Pinot, and Florian~Le Bronnec. 2026.
\newblock \href {https://arxiv.org/abs/2602.08660} {Equalized generative
  treatment: Matching f-divergences for fairness in generative models}.
\newblock \emph{Preprint}, arXiv:2602.08660.

\bibitem[{Verma and Rubin(2018)}]{verma-rubin-2018-fairness}
Sahil Verma and Julia Rubin. 2018.
\newblock \href {https://doi.org/10.1145/3194770.3194776} {Fairness definitions
  explained}.
\newblock In \emph{Proceedings of the International Workshop on Software
  Fairness}, {{FairWare}} '18, pages 1--7, New York, NY, USA. Association for
  Computing Machinery.

\bibitem[{Wang et~al.(2025)Wang, Adel, Lange, Liu, Nie, Str{\"o}tgen, and
  Schuetze}]{wang-etal-2025-lost-multilinguality}
Mingyang Wang, Heike Adel, Lukas Lange, Yihong Liu, Ercong Nie, Jannik
  Str{\"o}tgen, and Hinrich Schuetze. 2025.
\newblock \href {https://doi.org/10.18653/v1/2025.acl-long.253} {Lost in
  multilinguality: Dissecting cross-lingual factual inconsistency in
  transformer language models}.
\newblock In \emph{Proceedings of the 63rd Annual Meeting of the Association
  for Computational Linguistics (Volume 1: Long Papers)}, pages 5075--5094,
  Vienna, Austria. Association for Computational Linguistics.

\bibitem[{Weeber et~al.(2026)Weeber, Ceron, and
  Pad{\'o}}]{weeber-etal-2026-political}
Franziska Weeber, Tanise Ceron, and Sebastian Pad{\'o}. 2026.
\newblock \href {https://doi.org/10.48550/arXiv.2508.05553} {Do {{Political
  Opinions Transfer Between Western Languages}}? {{An Analysis}} of
  {{Unaligned}} and {{Aligned Multilingual LLMs}}}.
\newblock \emph{Preprint}, arXiv:2508.05553.

\bibitem[{Wendler et~al.(2024)Wendler, Veselovsky, Monea, and
  West}]{wendler-etal-2024-llamas}
Chris Wendler, Veniamin Veselovsky, Giovanni Monea, and Robert West. 2024.
\newblock \href {https://doi.org/10.18653/v1/2024.acl-long.820} {Do {{Llamas
  Work}} in {{English}}? {{On}} the {{Latent Language}} of {{Multilingual
  Transformers}}}.
\newblock In \emph{Proceedings of the 62nd {{Annual Meeting}} of the
  {{Association}} for {{Computational Linguistics}} ({{Volume}} 1: {{Long
  Papers}})}, pages 15366--15394, Bangkok, Thailand. Association for
  Computational Linguistics.

\bibitem[{Yang et~al.(2025)Yang, Li, Yang, Zhang, Hui, Zheng, Yu, Gao, Huang,
  Lv et~al.}]{yang-etal-2025-qwen3}
An~Yang, Anfeng Li, Baosong Yang, Beichen Zhang, Binyuan Hui, Bo~Zheng, Bowen
  Yu, Chang Gao, Chengen Huang, Chenxu Lv, and 1 others. 2025.
\newblock Qwen3 technical report.
\newblock \emph{arXiv preprint arXiv:2505.09388}.

\end{thebibliography}

\appendix

\section{Manifesto and Parlamint}%
\label{a:pm}

\begin{table}[!h]

    \centering
    \resizebox{\columnwidth}{!}{
    \begin{tabular}{lrr@{\extracolsep{.7em}}rr}
\toprule
\textbf{Language} & \multicolumn{2}{c}{\textbf{Parlamint}}   & \multicolumn{2}{c}{\textbf{Manifesto}}  \\
\cmidrule{2-3} \cmidrule{4-5}
 & \# Parties & \# Texts & \# Parties & \# Texts \\
\midrule
\textsc{bos} & 4 & 348 & 6 & 1245 \\
\textsc{bul} & 7 & 6477 & -- & -- \\
\textsc{cat} & 9 & 348 & 4 & 672 \\
\textsc{ces} & 7 & 8382 & -- & -- \\
\textsc{cnr} & -- & -- & 8 & 348 \\
\textsc{dan} & 12 & 4393 & 9 & 2426 \\
\textsc{deu} & 5 & 1265 & 22 & 9384 \\
\textsc{ell} & 6 & 819 & 7 & 683 \\
\textsc{eng} & 13 & 8884 & 49 & 21387 \\
\textsc{est} & 5 & 3710 & 5 & 1525 \\
\textsc{eus} & 2 & 164 & -- & -- \\
\textsc{fin} & -- & -- & 8 & 2745 \\
\textsc{fra} & 17 & 1850 & 7 & 4407 \\
\textsc{glg} & 3 & 1080 & -- & -- \\
\textsc{heb} & -- & -- & 14 & 2644 \\
\textsc{hrv} & 14 & 1376 & 10 & 1953 \\
\textsc{hun} & 10 & 2406 & 7 & 2991 \\
\textsc{hye} & -- & -- & 6 & 1313 \\
\textsc{isl} & -- & -- & 7 & 441 \\
\textsc{ita} & 8 & 964 & 9 & 2705 \\
\textsc{kat} & -- & -- & 7 & 802 \\
\textsc{kor} & -- & -- & 4 & 2216 \\
\textsc{lav} & 6 & 1284 & -- & -- \\
\textsc{lit} & -- & -- & 8 & 1339 \\
\textsc{nld} & 29 & 5541 & 24 & 18822 \\
\textsc{nno} & 6 & 180 & -- & -- \\
\textsc{nob} & 7 & 2070 & -- & -- \\
\textsc{pol} & 5 & 4266 & 7 & 1157 \\
\textsc{por} & 10 & 16677 & 15 & 4728 \\
\textsc{ron} & -- & -- & 4 & 797 \\
\textsc{slv} & 9 & 272 & 9 & 1138 \\
\textsc{spa} & 17 & 631 & 48 & 14908 \\
\textsc{srp} & -- & -- & 9 & 528 \\
\textsc{swe} & 8 & 2009 & 8 & 1696 \\
\textsc{tur} & 5 & 13541 & 6 & 3438 \\
\textsc{ukr} & 10 & 1820 & 2 & 126 \\
\bottomrule
\end{tabular}}

    \caption{Number of national parties and total number of texts for each language subset of the character-stratified subsets of Parlamint and Manifesto}
    \label{tab:statpm}
\end{table}

The number of parties and total number of texts for each language of Manifesto and Parlamint after character-stratification are reported in Table~\ref{tab:statpm}.

\begin{algorithm}[t]
\begin{algorithmic}
\State all\_chars $\gets$ number of characters for each text
\State bins $\gets$ 100 log-spaced bins 
\State min\_counts $\gets$ minimum number of texts for each bin over all parties 
\ForAll{party}
    \State party\_counts $\gets$ count texts per bin
    \State upscale $\gets \min(\frac{\mathrm{party\_counts}}{\mathrm{min\_counts}})$ 
    \ForAll{bin}
        \State $n \gets$ min\_counts[bin] $\times$ upscale
        \State sample $n$ texts from party[bin]
    \EndFor
\EndFor
\end{algorithmic}
\caption{Algorithm for character stratification of a given language subset}
\label{alg:strat}
\end{algorithm}

The algorithm for character stratification of a given language subset is given in Algorithm~\ref{alg:strat}.%

\section{Other metrics}
\label{a:metrics}
\subsection{Pretraining Fairness correlates with Downstream Performance}
Table~\ref{tab:cond_ppl_v_bpc+bpec} reports the Spearman $\rho$  correlation between source-conditioned BPC and BPEC and sBLEU- and COMET-based Borda  counts across political parties. 

Table~\ref{tab:trad_v_bpc+bpec} reports the Spearman $\rho$  correlation between unconditioned BPC and BPEC, and sBLEU- and COMET-based Borda  counts across political parties. 

\begin{table*}[t]
    \centering
    \begin{tabular}{rlrr@{\extracolsep{.7em}}rr}
    \toprule
       \textbf{Model}  & \textbf{Stage}& \multicolumn{2}{c}{\textbf{sBLEU}}   & \multicolumn{2}{c}{\textbf{COMET}} \\ 
       \cmidrule{3-4}
       \cmidrule{5-6}
        & & \textbf{BPC}$(t|s)$ & \textbf{BPEC}$(t|s)$ & \textbf{BPC}$(t|s)$ & \textbf{BPEC}$(t|s)$ \\ 
        \midrule
        Gemma-3-4B&Base & -0.619 & -0.476 & -0.786 & -0.714 \\ 
        Gemma-3-4B&Instruct & -0.524 & -0.310 & -0.690 & -0.571 \\ 
        Qwen3-4B&Base & -0.714 & -0.643 & -0.952 & -0.905 \\ 
        Qwen3-4B&Instruct & -0.643 & -0.452 & -0.905 & -0.762 \\ 
        Qwen3-8B&Base & -0.738 & -0.690 & -0.905 & -0.833 \\ 
        Qwen3-8B&Instruct & -0.690 & -0.619 & -0.833 & -0.810 \\ 
        Llama-3.1-8B&Base & -0.571 & -0.524 & -0.738 & -0.762 \\ 
        Llama-3.1-8B&Instruct & -0.667 & -0.524 & -0.833 & -0.762 \\ 
        Llama-3.1-70B&Base & -0.635 & -0.479 & -0.714 & -0.714 \\ 
        Llama-3.1-70B&Instruct & -0.695 & -0.515 & -0.810 & -0.762 \\ 
        \bottomrule
    \end{tabular}%
    \caption{Spearman $\rho$ shows a negative correlation between source-conditioned BPC ($\downarrow$) and BPEC ($\downarrow$) and sBLEU- and COMET-based Borda counts ($\uparrow$) across political parties, both for the source-conditioned BPC and BPEC of base LLMs  and of IT-LLMs.}
    \label{tab:cond_ppl_v_bpc+bpec}

    \centering
    \begin{tabular}{rlrr@{\extracolsep{.7em}}rr}
    \\ \\
    \toprule
       \textbf{Model}  & \textbf{Stage}& \multicolumn{2}{c}{\textbf{sBLEU}}   & \multicolumn{2}{c}{\textbf{COMET}} \\ 
       \cmidrule{3-4}
       \cmidrule{5-6}
        & & \textbf{BPC}$(t)$ & \textbf{BPEC}$(t)$ & \textbf{BPC}$(t)$ & \textbf{BPEC}$(t)$ \\ 
        \midrule
        Gemma-3-4B&Base & -0.357 & -0.238 & -0.548 & -0.524 \\ 
        Gemma-3-4B&Instruct & -0.262 & -0.310 & -0.500 & -0.571 \\ 
        Qwen3-4B&Base & -0.643 & -0.452 & -0.905 & -0.762 \\ 
        Qwen3-4B&Instruct & -0.524 & -0.452 & -0.810 & -0.762 \\ 
        Qwen3-8B&Base & -0.690 & -0.476 & -0.833 & -0.643 \\ 
        Qwen3-8B&Instruct & -0.571 & -0.476 & -0.714 & -0.643 \\ 
        Llama-3.1-8B&Base & -0.429 & -0.310 & -0.619 & -0.595 \\ 
        Llama-3.1-8B&Instruct & -0.310 & -0.310 & -0.548 & -0.595 \\ 
        Llama-3.1-70B&Base & -0.527 & -0.335 & -0.524 & -0.595 \\ 
        Llama-3.1-70B&Instruct & -0.491 & -0.335 & -0.595 & -0.595 \\ 
        \bottomrule
    \end{tabular}
    \caption{Spearman $\rho$ shows a negative correlation between BPC ($\downarrow$) and BPEC ($\downarrow$) and sBLEU- and COMET-based Borda counts ($\uparrow$) across political parties, both for the BPC and BPEC of base LLMs  and of IT-LLMs. BPC and BPEC are computed on the target text $t$, without context, especially without the source text.}
    \label{tab:trad_v_bpc+bpec}
\end{table*}

\subsection{Pretraining Fairness on a Parallel Dataset}\label{sa:pt_metrics}
\begin{table*}[t]
    \centering
    \begin{tabular}{rlrrrrrrrr}
\toprule
Model & Stage & NGL & S\&D & EFA & ALDE & EPP & ECR & EFD & NA \\
\midrule
Gemma-3-4B & Base & \cellcolor{green!17} 1.03 & \cellcolor{green!21} 0.99 & \cellcolor{red!30} 1.06 & \cellcolor{red!18} 1.04 & \cellcolor{green!25} 0.99 & \cellcolor{red!22} 1.05 & \cellcolor{green!30} 0.99 & \cellcolor{red!26} 1.06\\ 

Gemma-3-4B & Instruct & \cellcolor{green!17} 1.43 & \cellcolor{green!21} 1.39 & \cellcolor{red!26} 1.48 & \cellcolor{red!18} 1.44 & \cellcolor{green!30} 1.38 & \cellcolor{red!22} 1.46 & \cellcolor{green!25} 1.39 & \cellcolor{red!30} 1.49\\ 

Qwen3-4B & Base & \cellcolor{red!18} 1.09 & \cellcolor{green!25} 1.04 & \cellcolor{red!26} 1.11 & \cellcolor{green!17} 1.09 & \cellcolor{green!30} 1.04 & \cellcolor{red!22} 1.11 & \cellcolor{green!21} 1.05 & \cellcolor{red!30} 1.12\\ 

Qwen3-4B & Instruct & \cellcolor{red!18} 1.30 & \cellcolor{green!25} 1.25 & \cellcolor{red!26} 1.33 & \cellcolor{green!17} 1.30 & \cellcolor{green!30} 1.25 & \cellcolor{red!22} 1.32 & \cellcolor{green!21} 1.26 & \cellcolor{red!30} 1.35\\ 

Qwen3-8B & Base & \cellcolor{red!18} 1.02 & \cellcolor{green!25} 0.97 & \cellcolor{red!26} 1.04 & \cellcolor{green!17} 1.01 & \cellcolor{green!30} 0.97 & \cellcolor{red!22} 1.03 & \cellcolor{green!21} 0.98 & \cellcolor{red!30} 1.05\\ 

Qwen3-8B & Instruct & \cellcolor{red!18} 1.17 & \cellcolor{green!25} 1.12 & \cellcolor{red!26} 1.20 & \cellcolor{green!17} 1.17 & \cellcolor{green!30} 1.12 & \cellcolor{red!22} 1.19 & \cellcolor{green!21} 1.13 & \cellcolor{red!30} 1.21\\ 

Llama-3.1-8B & Base & \cellcolor{green!17} 1.10 & \cellcolor{green!21} 1.06 & \cellcolor{red!26} 1.13 & \cellcolor{red!18} 1.11 & \cellcolor{green!25} 1.06 & \cellcolor{red!22} 1.12 & \cellcolor{green!30} 1.06 & \cellcolor{red!30} 1.14\\ 

Llama-3.1-8B & Instruct & \cellcolor{green!17} 1.15 & \cellcolor{green!21} 1.12 & \cellcolor{red!26} 1.19 & \cellcolor{red!18} 1.17 & \cellcolor{green!25} 1.12 & \cellcolor{red!22} 1.18 & \cellcolor{green!30} 1.12 & \cellcolor{red!30} 1.20\\ 

Llama-3.1-70B & Base & \cellcolor{green!17} 0.98 & \cellcolor{green!30} 0.94 & \cellcolor{red!26} 1.01 & \cellcolor{red!18} 1.00 & \cellcolor{green!21} 0.95 & \cellcolor{red!22} 1.01 & \cellcolor{green!25} 0.95 & \cellcolor{red!30} 1.02\\ 

Llama-3.1-70B & Instruct & \cellcolor{green!17} 1.03 & \cellcolor{green!21} 0.99 & \cellcolor{red!26} 1.06 & \cellcolor{red!18} 1.05 & \cellcolor{green!25} 0.99 & \cellcolor{red!22} 1.06 & \cellcolor{green!30} 0.99 & \cellcolor{red!30} 1.07\\ 
\bottomrule
\end{tabular}
    \caption{BPC ($\downarrow$) of base and IT-LLMs for the eight political parties of 21-EuroParl. BPC is computed on the target text $t$, without context, especially without the source text.}
    \label{tab:BPC}

    \centering
    \begin{tabular}{rlrrrrrrrr}
    \\ \\
\toprule
Model & Stage& NGL & SD & EFA & ALDE & EPP & ECR & EFD & NA\\

\midrule

Gemma-3-4B & Base & \cellcolor{green!17} 1.05 & \cellcolor{green!25} 1.02 & \cellcolor{red!30} 1.09 & \cellcolor{red!18} 1.07 & \cellcolor{green!30} 1.01 & \cellcolor{red!26} 1.08 & \cellcolor{green!21} 1.03 & \cellcolor{red!22} 1.08\\ 

Gemma-3-4B & Instruct & \cellcolor{green!17} 1.45 & \cellcolor{green!21} 1.44 & \cellcolor{red!30} 1.53 & \cellcolor{red!18} 1.49 & \cellcolor{green!30} 1.41 & \cellcolor{red!22} 1.50 & \cellcolor{green!25} 1.44 & \cellcolor{red!26} 1.52\\ 

Qwen3-4B & Base & \cellcolor{green!17} 1.12 & \cellcolor{green!25} 1.09 & \cellcolor{red!30} 1.16 & \cellcolor{red!18} 1.13 & \cellcolor{green!30} 1.07 & \cellcolor{red!26} 1.15 & \cellcolor{green!21} 1.10 & \cellcolor{red!22} 1.15\\ 

Qwen3-4B & Instruct & \cellcolor{green!17} 1.34 & \cellcolor{green!25} 1.31 & \cellcolor{red!30} 1.39 & \cellcolor{red!18} 1.36 & \cellcolor{green!30} 1.29 & \cellcolor{red!26} 1.38 & \cellcolor{green!21} 1.32 & \cellcolor{red!22} 1.38\\ 

Qwen3-8B & Base & \cellcolor{green!17} 1.04 & \cellcolor{green!25} 1.01 & \cellcolor{red!30} 1.08 & \cellcolor{red!18} 1.06 & \cellcolor{green!30} 1.00 & \cellcolor{red!26} 1.08 & \cellcolor{green!21} 1.02 & \cellcolor{red!22} 1.07\\ 

Qwen3-8B & Instruct & \cellcolor{green!17} 1.20 & \cellcolor{green!25} 1.17 & \cellcolor{red!30} 1.25 & \cellcolor{red!18} 1.22 & \cellcolor{green!30} 1.15 & \cellcolor{red!26} 1.24 & \cellcolor{green!21} 1.18 & \cellcolor{red!22} 1.24\\ 

Llama-3.1-8B & Base & \cellcolor{green!17} 1.12 & \cellcolor{green!21} 1.10 & \cellcolor{red!30} 1.17 & \cellcolor{red!18} 1.15 & \cellcolor{green!30} 1.08 & \cellcolor{red!26} 1.16 & \cellcolor{green!25} 1.10 & \cellcolor{red!22} 1.16\\ 

Llama-3.1-8B & Instruct & \cellcolor{green!17} 1.17 & \cellcolor{green!21} 1.16 & \cellcolor{red!30} 1.23 & \cellcolor{red!18} 1.21 & \cellcolor{green!30} 1.14 & \cellcolor{red!26} 1.22 & \cellcolor{green!25} 1.16 & \cellcolor{red!22} 1.22\\ 

Llama-3.1-70B & Base & \cellcolor{green!17} 1.00 & \cellcolor{green!21} 0.98 & \cellcolor{red!30} 1.05 & \cellcolor{red!18} 1.03 & \cellcolor{green!30} 0.97 & \cellcolor{red!26} 1.05 & \cellcolor{green!25} 0.98 & \cellcolor{red!22} 1.04\\ 

Llama-3.1-70B & Instruct & \cellcolor{green!17} 1.05 & \cellcolor{green!21} 1.03 & \cellcolor{red!30} 1.10 & \cellcolor{red!18} 1.08 & \cellcolor{green!30} 1.01 & \cellcolor{red!26} 1.09 & \cellcolor{green!25} 1.03 & \cellcolor{red!22} 1.09\\ 

\bottomrule
\end{tabular}
    \caption{BPEC ($\downarrow$) of base and IT-LLMs for the eight political parties of 21-EuroParl. BPEC is computed on the target text $t$, without context, especially without the source text.}
    \label{tab:BPEC}
\end{table*}

Tables~\ref{tab:BPC} and \ref{tab:BPEC} respectively report the BPC and the BPEC of base and IT-LLMs for the eight political parties of 21-EuroParl. %

\clearpage
\listoftodos

\end{document}